\documentclass[10pt, conference]{IEEEtran} 
\IEEEoverridecommandlockouts

\usepackage[utf8]{inputenc} 
\usepackage[T1]{fontenc}    
\usepackage{hyperref}       
\usepackage{url}            
\usepackage{booktabs}       
\usepackage{amsfonts}       
\usepackage{nicefrac}       
\usepackage{microtype}      
\usepackage{xcolor}         
\usepackage{amsmath,amssymb,multicol,latexsym}
\usepackage{pgfplots,pgfplotstable}
\pgfplotsset{compat=1.16}
\usepackage{tikz}
\usepackage{hyperref}
\usepackage{cleveref}

\DeclareMathAlphabet{\mathcal}{OMS}{cmsy}{m}{n}

\usepackage{url}
\usepackage{graphicx}
\usepackage{cite}
\usepackage{flushend} 

\usepackage[colorinlistoftodos,prependcaption,textsize=small]{todonotes}
\usepackage{regexpatch}

\makeatletter

\graphicspath{{figures/}}

\usepackage{tcolorbox}
\tcbuselibrary{skins}
\newcommand{\gradientbox}[3]{{\raisebox{-1pt}{\tcbox[enhanced, top=1pt,bottom=1pt,left=1pt,right=1pt,middle=0pt, boxrule=0pt, arc=0pt, boxsep=0pt, interior style={left color=#1,right color=#2}]{#3}}}}

\title{\LARGE \bf Traffic Scene Similarity: a Graph-based Contrastive Learning Approach}

\author{
Maximilian~Zipfl$^{1,2}$,
Moritz~Jarosch$^{2}$, 
and J.~Marius~Zöllner$^{1,2}$
\thanks{$^{1}$FZI Research Center for Information Technology, Karlsruhe, Germany
{\tt\small zipfl@fzi.de}}%
\thanks{$^{2}$Karlsruhe Institute of Technology, Karlsruhe, Germany}%
}

\date{May 2023}

\makeatletter
\def\ps@IEEEtitlepagestyle{%
  \def\@oddfoot{\mycopyrightnotice}%
  \def\@evenfoot{}%
}
\def\mycopyrightnotice{%
  \begin{minipage}{\textwidth}
  \centering \scriptsize
  Copyright~\copyright~2023 IEEE. Personal use of this material is permitted. Permission from IEEE must be obtained for all other uses, in any current or future media, including\\
  reprinting/republishing this material for advertising or promotional purposes, creating new collective works, for resale or redistribution to servers or lists, or reuse of any copyrighted component of this work in other works.
  \end{minipage}
}
\makeatother

\begin{document}

\maketitle

\begin{abstract}
Ensuring validation for highly automated driving poses significant obstacles to the widespread adoption of highly automated vehicles. Scenario-based testing offers a potential solution by reducing the homologation effort required for these systems. However, a crucial prerequisite, yet unresolved, is the definition and reduction of the test space to a finite number of scenarios.

To tackle this challenge, we propose an extension to a contrastive learning approach utilizing graphs to construct a meaningful embedding space. Our approach demonstrates the continuous mapping of scenes using scene-specific features and the formation of thematically similar clusters based on the resulting embeddings. 
Based on the found clusters, similar scenes could be identified in the subsequent test process, which can lead to a reduction in redundant test runs.
\end{abstract}


\section{Introduction}
The validation of highly automated driving systems remains a major challenge for their widespread integration into public road traffic. The vast operational design domain (ODD) and the resulting large number of test possibilities make the application of statistical test methods economically unfeasible. Consequently, alternative verification strategies are essential to address this issue \cite{wachenfeld_release_2016}. 
Scenario-based testing as a validation methodology promises to solve several challenges \cite{riedmaier_survey_2020}. 
The utilization of authentic data enables the identification of plausible scenarios that possess the potential to trigger errors. Additionally, the test scenarios generated from the data hold promise for achieving robust test coverage.
One challenge in data-driven scenario generation is the classification of scenarios and scenes. 
The initial step in the scenario-based testing process involves defining the test space, which refers to the entire scope of potential test scenarios. This is achieved by dividing the test space into specific areas that can be effectively tested using scenario-based testing methods.
Typically, the prevailing approach involves the manual annotation of scenarios and scenes that are subsequently clustered and grouped together.

One of the most important prerequisites for automating the clustering process is to identify a descriptive model that can cover a broad range of possible scenarios, including edge cases, while still maintaining sufficient granularity to distinguish between distinct situations.

The focus of this paper is on the transformation of a traffic scene, described by a graph, into a projection that allows traffic scenes to be grouped and categorised according to their similarity. This work builds heavily on our previous publications \cite{zipfl_self_2022} in this field and can be seen as a continuation of our earlier work. 
We were able to show that with the help of a Graph Neural Network (GNN) contrastive learning approach, similar traffic scenes represented as graphs can be grouped and that the procedure is applicable in principle.
In this paper, we address the challenge of validating the similarity of unlabelled traffic scenes.

\begin{figure}[t]
    \centering
    \def\svgwidth{0.95\columnwidth}
\begingroup%
  \makeatletter%
  \providecommand\color[2][]{%
    \errmessage{(Inkscape) Color is used for the text in Inkscape, but the package 'color.sty' is not loaded}%
    \renewcommand\color[2][]{}%
  }%
  \providecommand\transparent[1]{%
    \errmessage{(Inkscape) Transparency is used (non-zero) for the text in Inkscape, but the package 'transparent.sty' is not loaded}%
    \renewcommand\transparent[1]{}%
  }%
  \providecommand\rotatebox[2]{#2}%
  \newcommand*\fsize{\dimexpr\f@size pt\relax}%
  \newcommand*\lineheight[1]{\fontsize{\fsize}{#1\fsize}\selectfont}%
  \ifx\svgwidth\undefined%
    \setlength{\unitlength}{317.68157238bp}%
    \ifx\svgscale\undefined%
      \relax%
    \else%
      \setlength{\unitlength}{\unitlength * \real{\svgscale}}%
    \fi%
  \else%
    \setlength{\unitlength}{\svgwidth}%
  \fi%
  \global\let\svgwidth\undefined%
  \global\let\svgscale\undefined%
  \makeatother%
  \begin{picture}(1,0.59202798)%
    \lineheight{1}%
    \setlength\tabcolsep{0pt}%
    \put(0,0){\includegraphics[width=\unitlength,page=1]{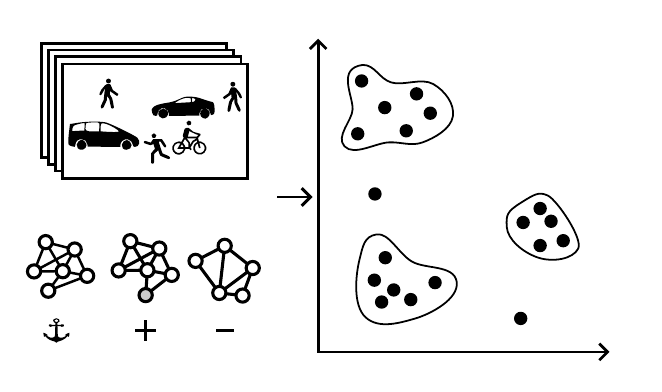}}%
    \put(0.7362715,0.50448353){\color[rgb]{0,0,0}\makebox(0,0)[t]{\lineheight{1.25}\smash{\begin{tabular}[t]{c}Type A\end{tabular}}}}%
    \put(0,0){\includegraphics[width=\unitlength,page=2]{top_right.pdf}}%
    \put(0.84088001,0.36408767){\color[rgb]{0,0,0}\makebox(0,0)[t]{\lineheight{1.25}\smash{\begin{tabular}[t]{c}Type C\end{tabular}}}}%
    \put(0.67020271,0.2365385){\color[rgb]{0,0,0}\makebox(0,0)[t]{\lineheight{1.25}\smash{\begin{tabular}[t]{c}Type B\end{tabular}}}}%
    \put(0,0){\includegraphics[width=\unitlength,page=3]{top_right.pdf}}%
  \end{picture}%
\endgroup%

    \caption{Schematic representation of our approach: Generating a sample triplet for self-supervised clustering of traffic scenes.}
    \label{fig:approach_topright}
\end{figure}

This work adds to this discussion in two ways. Firstly, we utilized a larger and more advanced model, which was trained on an expanded dataset that included diverse street types from multiple countries. Secondly, we conducted a thorough and detailed analysis of our results, specifically with regard to the encoding space and the clustering outcomes.

This work makes the following key contributions:
\begin{itemize}
    \item[(i)] We propose a method to map traffic scenes, represented as graphs, onto a low-dimensional embedding space.
    \item[(ii)] We conduct a comprehensive quantitative analysis of the embedding space, focusing on the information content of the mapped traffic scenes.
    \item[(iii)] The resulting embedding space facilitates the clustering of traffic scenes with similar themes. 
\end{itemize}
The rest of the work is structured as follows.
\Cref{sec:related_work} gives a short introduction to similar approaches in the context of scenario or scene clustering. 
\Cref{sec:methodology} describes the principles of this approach and \Cref{sec:implementation} gives insight to the developed architecture and the training process.
After that, \Cref{sec:analysis} presents an evaluation and compares its performance on a motion dataset. 
Finally, we conclude our paper in \Cref{sec:conclusion}.

\vspace*{1cm}

\section{Related Work}
\label{sec:related_work}
In this section, we present related work on data-based clustering of unlabelled traffic scenarios and traffic scenes, and we briefly introduce the graph description model used in this work to depict traffic scenes.

\subsection{Clustering of Traffic Scenes and Scenarios}
In recent years, a number of works have been published that address the clustering or categorisation of traffic scenes.
We have decided to divide the recent work into two categories. Firstly, feature based approaches, where different attributes and time series of features are compared against each other. On the other hand, manoeuvre based approaches take this further, using mainly the trajectories and movement patterns of vehicles as input to describe the traffic scene.

\subsubsection{Feature Based Approaches}
The work of Kruber et al. \cite{kruber_unsupervised_2019} uses features of a highway scenario in regard to an ego vehicle (distances to other traffic participants, velocities, ...) to group similar traffic scenarios using a random forest approach.

In \cite{kerber_clustering_2020} traffic scenes are compared over time by means of an eight-car neighbourhood model to calculate the similarity of highway scenarios. 
Scenarios are compared in pairs based on the distances of other traffic participants in the eight areas around the ego vehicle. 
Only scenarios in the same location and with the same number of traffic participants can be meaningfully compared.
The resulting clusters contain similar manoeuvres of both the ego and the surrounding vehicles. 
Hauer et al. \cite{hauer_clustering_2020} uses a similar concept where the time series of multiple features are compared by using Dynamic Time Warping (DTW). To reduce the dimensionality a Principal Component Analysis (PCA) is used to project the data to a clustering space.

\subsubsection{Manoeuvre Based Approaches}
Bernhard et al. \cite{bernhard_optimizing_2021} use the bag-of-words method to convert trajectories through spatially divided clusters to histograms. The affinity of the scenarios are described by the similarity of the corresponding histograms.
Another approach by Ries et al. \cite{ries_trajectory-based_2021} uses DTW to compare trajectories of traffic participant. Scenarios are similar and can be clustered once two scenarios exist with the same traffic participant types and similar trajectories.

Another approach by Harmening et al. \cite{harmening_deep_2020} presents two autoencoder-based models that reconstruct traffic scenes based on a grid map and traffic scenarios on temporally related position and speed features.
In each case, the embedding space is used for clustering. 

In the paper by Wurst et al. \cite{wurst_expert-lasts_2022} the authors present a machine learning based method that allows the grouping of similar scenarios and novelty detection at the same time. The road geometry is used on the basis of a birds-eye image and road topology by means of a graph to mine samples for a contrastive learning approach. 

The survey by Bian et al. \cite{bian_survey_2018} provides an overall insight into the possibilities of comparing trajectories and gives a broad overview of recent work on clustering vehicle trajectories.

\subsection{Semantic Scene Graph}
\label{sec:semantic_scene_graph}

\begin{figure}[htbp]
  \centering
  \def\svgwidth{\columnwidth}
  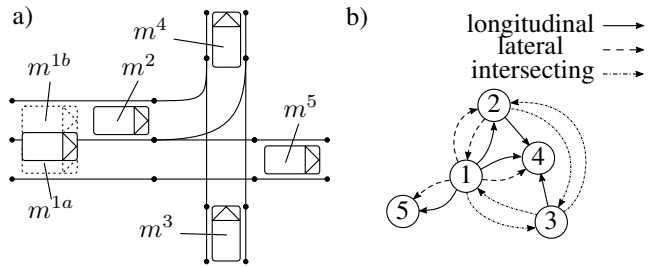
  \caption{Exemplary traffic scene with five vehicles and six corresponding projection identities $m$ (a). The resulting scene graph, where each traffic participant is represented by a node and the relations between its projection identities as edges (b)  \cite{zipfl_towards_2021}.}
  \label{fig:relation_types}
\end{figure}

To employ machine learning methods in processing traffic scenes, it is necessary to convert them into a machine-readable format. In this regard, we utilize the Semantic Scene Graph (SSG) introduced in \cite{zipfl_towards_2021}. This approach strategically disregards the absolute positions of traffic participants and emphasizes their interactions. Based on the topology of the road network and the relative positions of traffic participants, a heterogeneous graph is generated, which encode the relationships between traffic participants.
Each graph $G = (V,E)$ is defined by the nodes $v \in V$ which represent the traffic participants and the edges $e^{ij} \in E$ between two nodes ($v^i,v^j$), which represent their relations. 
These edges fall into three categories: longitudinal (for traffic participants on the same lane), lateral (for traffic participants on two parallel lanes), and intersecting (for traffic participants at intersections or motorway ramps). Node attributes store velocity information and traffic participant classification, while edge attributes store information about edge type and the distance in Frenet space along the lanes, between the two regarding traffic participants. This abstraction allows traffic scenes to be described independently of road geometry and focusing solely on the traffic constellation, thereby enhancing comparability.
In order to preserve the information in the graph, that vehicles are not exactly in the middle of a lane, projection identities are introduced that represent the certainty (probability) that a vehicle is in a given lane.
In \Cref{fig:relation_types}a you can see an exemplary traffic scene with five vehicles and a simple road network. The corresponding SGG is shown in \Cref{fig:relation_types}b.
We refer the reader to the paper by Zipfl et al. \cite{zipfl_towards_2021} for further details on the SGG and its implementation.

\section{Methodology}
\label{sec:methodology}

The objective of our methodology is to minimize the distance between the embeddings $s_0, s_+$ of two similar graphs, $G_0$ and $G_+$, in the embedding space $S$.
In addition, it is important that the embedding $s_-$ of a dissimilar graph $G_-$ be as distant from $s_0$ as feasible.
We achieve this by implementing a contrastive learning approach to address this graph representation problem.

\subsection{Graph Augmentation}
The object list data from a motion dataset serves as the foundation for our methodology.
Our approach involves modifying the traffic scene by means of the object list, as opposed to directly altering the graph features. This offers the advantage of maintaining the consistency and realism of the traffic scenes generated. By contrast, random additions, deletions, or modifications of edges and nodes within the graph representation might lead to traffic scenes that are improbable or impossible in reality. 
During the augmentation process, traffic participants are chosen at random and subjected to modifications of their speed and position that conform to a standard distribution.
For further information on the traffic scene augmentation, please refer to the previous paper \cite{zipfl_self_2022}.

Afterward, both the augmented and original scenes are converted into a graph representation using the SGG (see \Cref{sec:semantic_scene_graph}).

\subsection{Model Architecture}

\begin{figure}[htbp]
    \centering
    \includegraphics[width=1.0\linewidth]{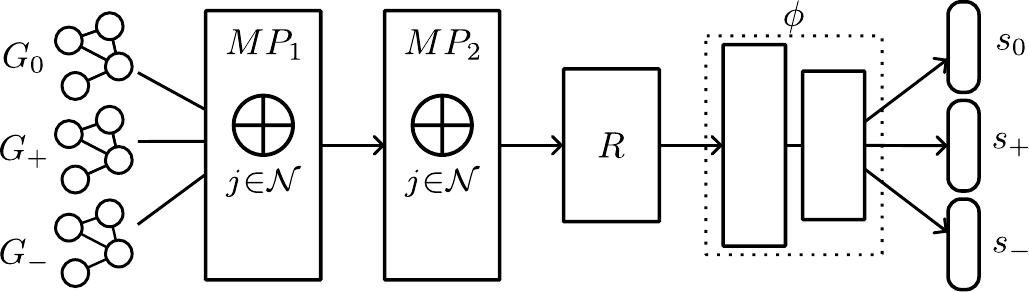}
    \caption{Contrastive learning model mapping graph-triplets onto the embedding space\cite{zipfl_self_2022}.}
    \label{fig:representation_pipeline}
\end{figure}

The model architecture used for encoding a traffic scene graph into an embedding is a Graph Neural Network with two layers, a readout function in form of a summation over all hidden node states and a projection head, as seen in \Cref{fig:representation_pipeline}.

The principle of the message passing approach we use can be explained with the following formula:

 \begin{align}
     v^i_k = \gamma_k(v^i_{k-1}, \bigoplus_{j \in \mathcal{N}(i)}(m_k(v^i_{k-1}, v^j_{k-1}, e^{ji}))).
     \label{eq:message_passing}
 \end{align}

$k$ represents the layer of the GNN, while $i$ specifies a specific node. The first step in calculating the new state for $v^i_k$ is to generate messages along all incoming edges to the node. This message function $m_k$ combines the features of the incoming $i$ and outgoing $j$ nodes, as well as the edge features $e^{ji}$. Afterwards, an aggregation function $\bigoplus$ is applied to reduce the arbitrary amount of messages to one of fixed dimensionality. The resulting vector is then fed through the update function $\gamma_k$, along with the incoming node features, resulting in the final new node state $v^i_k$.

The first GNN layer uses both node and edge features, while the second one only operates on the hidden node states. Instead of single linear layers in the message passing and update function of the GNN layers, two layer MLPs (Multi Layer Perceptrons) are used. Nonlinearity is added through LeakyReLU activation functions. Regularization is done through dropout in the GNN layers.
The projection head is a three layer MLP.
All hidden node states have a dimensionality of 60 and the embeddings $s \in S$ are 12 dimensional ($\mathcal{D} = 12)$.

\section{Implementation and Training}
\label{sec:implementation}

\subsection{Dataset}
We start our processing pipeline by utilizing traffic scenes extracted from INTERACTION motion dataset \cite{zhan_interaction_2019}. The dataset was considered suitable for our study due to its diverse range of traffic scenarios that include intersections and other road sections from multiple countries. Additionally, assigning individual scenes to specific locations provides a useful pseudo label for validating our approach. The dataset was recorded via a drone, allowing for consistent observations of the same locations, across each sequence. Apart from object tracks, the dataset also includes corresponding HD maps of the roads. Each object in the dataset is defined by its pose in Cartesian space, classification, and velocity, sampled at a rate of 10 Hz.
\Cref{tab:dataset} displays an extract from the dataset under investigation, featuring 9306 scenes categorized according to their respective road types.
\begin{table}[tbp]
\caption{Composition of the dataset used.}
\label{tab:dataset}
\centering
\begin{tabular}{@{}lccc@{}}
 \toprule
    Road Type & Country & Name & Num. Scenes \\
 \midrule
    Roundabout      & GER & OF & 1822 \\ 
    Merging         & GER & MT & 863 \\
    Intersection    & USA & EP0 & 1651 \\
    Intersection    & USA & EP1 & 1349\\
    Roundabout      & USA & EP & 1769\\ 
    Roundabout      & USA & SR & 1852\\ 
  \textbf{Total}  & & & \textbf{9306} \\
\bottomrule
\end{tabular}
\end{table}

The dataset is first evenly split into a holdout set used for clustering and evaluating the results (20\%) and a train/test (80\%) set. The train test set is then again split into train (80\%) and validation (20\%) for parameter tuning.

The graphs used have both node and edge features. For node features, the object classification and its velocity are used. We have excluded the directional information of vehicle velocities or orientations in our approach, as the aim is to generalize traffic scenes without relying on their absolute position. 
Our emphasis is on the interdependent relationships between the different traffic participants, rather than their absolute, individual characteristics. \Cref{tab:features} provides an overview of the edge and node features utilized in our approach.

The original implementation of the SGG includes parallel edges connecting two traffic participants with multiple projection identities. However, in this study, we propose a modification whereby the parallel edges are consolidated into a single primary edge. Furthermore, we combine the certainties of individual types and represent them as a three-valued attribute vector, in the range of $[0,1]$ on the merged edge.
For more detailed information on the features, please refer to the scene graph paper \cite{zipfl_towards_2021}.

\begin{table}[htb]
\centering
\caption{Description of the node and edge features utilized for training.}
\label{tab:features}
\begin{tabular}{@{}p{2cm}p{5.3cm}@{}}
\toprule
    Node Feature &\\
 \midrule
  velocity & The norm of the objects' velocity vector.\vspace{0.5ex}\\
  object class & One-hot encoding of the object type [car, truck, pedestrian, bike]\\
 \midrule
    Edge Feature &\\
 \midrule
  lat/lon/int certainty & The normalized level of certainty regarding the longitudinal, lateral, or intersecting relationship between the origin and target objects.\vspace{0.5ex}\\
  path distance & Frenet space distance is calculated when there is a longitudinal/lateral relationship (certainty $>0$) between the origin and target objects.\vspace{0.5ex}\\
  int path distance & If intersecting certainty $>0$, the distance from the origin object to the intersection of its own lane and the target object's lane is calculated.\vspace{0.5ex}\\
  origin centerline distance & The distance to the centerline of the lane associated with the origin object's projection identity having the highest certainty value (either lateral or longitudinal).\vspace{0.5ex}\\
  target centerline distance & The same as \emph{origin centerline distance} but for the target object.\vspace{0.5ex}\\
  int origin centerline distance & Refer to \emph{origin centerline distance}, which is computed solely when there is a positive intersecting certainty value.\vspace{0.5ex}\\
  int target centerline distance &  The same as \emph{int origin centerline distance} but for the target object.\vspace{0.5ex}\\
 \bottomrule
\end{tabular}
\end{table}

\subsection{Training procedure}

Learning is conducted using the triplet loss function $\mathcal{L}$ from \Cref{eq:triplet}.
\begin{align}
    \label{eq:triplet}
    \mathcal{L}(s_0, s_+, s_-) = max\left(d(s_0, s_+) - d(s_0, s_-) + M, 0\right)
\end{align}

Our approach involves utilizing graph encodings $s_0$, $s_+$, and $s_-$ ($s_0, s_+, s_- \in S$), where $s_0$ serves as the encoding for the original graph, also referred to as the anchor. The augmented encoding, known as the positive sample, is represented by $s_+$, while $s_-$ corresponds to a differing, randomly sampled graph that serves as the negative example. Negative samples in our study are uniformly sampled from the remainder of the batch. The distance metric function $d(\cdot)$ can be any suitable method, such as the Euclidean distance or cosine similarity, and is used to distance negative samples by a specified margin $M$.
In our application we got the best results with the Euclidean distance (\Cref{eq:distance_formula}), where $\mathcal{D}$ defines the dimensionality of the $S$.
\begin{align}
    \label{eq:distance_formula}
    d(s_0,s_+) = ||s_+ - s_0||_\mathcal{D} 
\end{align}

Our training conditions can be seen in \Cref{tab:training_parameters}.
Negative sampling was conducted uniformly over the batch, excluding the positive sample.

\begin{table}[]
\centering
\caption{The training hyperparameters}
\label{tab:training_parameters}
\begin{tabular}{@{}lc@{}}
 \toprule
    Parameter & Value \\
 \midrule
    Learning rate   & 0.001 \\ 
    Margin $M$      & 0.5 \\
    Embedding dimensionality $\mathcal{D}$ & 12\\
    batch size      & 400 \\
    Epochs          & 400\\
    Optimizer       & ADAM\\
 \bottomrule
\end{tabular}
\end{table}

\section{Analysis}
\label{sec:analysis}

\subsection{Embedding Space}

Our initial goal is to determine whether the trained model effectively embed similar traffic scenes in the embedding space $S$. In order to assess this, we conducted experiments on the holdout set, using previously unseen data. To test the network's discriminative ability, positive samples $s_+$ should be closer in distance to the anchor $s_0$ than negative samples $s_-$, thus correctly classifying them.
We define this assumption by \Cref{eq:assumption}.
\begin{equation}
    \label{eq:assumption}
     d(s_0,s_+) < d(s_0,s_-)
\end{equation}
The accuracy of the trained model for various partial datasets is presented in \Cref{tab:classification_distances}. $\bar{d}$ is the average distance measured for each location. The overall accuracy of the model for the test dataset is approximately 99.3\%. For the locations MT and SR, all samples of the holdout set are classified correctly. Combined with the consistent distances between anchor, positive and negative samples, we find that our model has generalized its discriminative properties.

\begin{table}
\centering
\caption{Classification accuracy and average distances between anchor, positive and negative sample}
\label{tab:classification_distances}
\begin{tabular}{@{}l c c c c c@{}}
 \toprule
    Street Type & Location & Accuracy & $\bar{d}(s_0,s_+)$ & $\bar{d}(s_0,s_-)$\\ 
 \midrule
    Roundabout &  OF & 0.989 & 0.385 & 2.921\\ 
    Merging &  MT & 1.0 & 0.471 & 3.079\\
    Intersection & EP0 & 0.99 & 0.439 & 2.628\\
    Intersection & EP1 & 0.993 & 0.458 & 3.057\\
    Roundabout & EP & 0.994 & 0.466 & 2.944\\ 
    Roundabout & SR & 1.0 & 0.439 & 3.108\\ 
    \textbf{Total} & & \textbf{0.993} & \textbf{0.439} &\textbf{2.927}\\ 
    \bottomrule
\end{tabular}

\end{table}

The task of labelling traffic scenes presents a challenging problem in the field of research. While it may be feasible to establish a concise set of rule-based labels, such as "following" or "overtaking," for scenarios involving only a small number of vehicles, the complexity of our traffic scenes far surpasses this scale, often comprising several dozen vehicles at maximum. Consequently, attempting to manually label such intricate interactions or relying on rule-based methods would prove to be unproductive. As a result, an alternative approach becomes imperative for the purpose of validation.

Our approach utilizes features derived from the graph structure itself. These features, although not as discerning as categorical labels, offer valuable aggregated information regarding the defining characteristics of a given traffic scene.

We use those features to evaluate how well our embedding space captures semantic information of the traffic scenes. Again using the holdout set, we craft multiple features and use a small four layer MLP (30 hidden dimensions) to regress from $S$ to the corresponding feature. For this, the holdout set is split again into a train/test (80/20) and we report the regression errors for the test set in \Cref{tab:hand_crafted_features_regression}.
 Training was conducted over 2500 Epochs, with a learning rate of 0.001, 10\% dropout, optimized using ADAM. The model was trained on Mean Squared Error (MSE). To bring the resulting errors into context, we include the Mean Absolute Error (MAE) of the prediction in combination with the Mean, Standard Deviation (Std), Min and Max of the regressed feature in the test dataset.

\begin{table}[t]
\centering
\caption{Regression performance of the embedding space onto handcrafted features.}
\label{tab:hand_crafted_features_regression}
\begin{tabular}{@{}l c c c c c c@{}}
    \toprule
    Regr. Var. & MSE & MAE & Mean & Std & Min & Max \\ \midrule
    $|E_{lon}|$ & 0.572 & 0.553 & 1.937 & 1.37 & 0.0 & 8.409 \\ 
    $|E_{lat}|$ & 0.885 & 0.218 & 0.243 & 0.388 & 0.0 & 2.067 \\ 
    $|E_{int}|$ & 0.367 & 0.588 & 3.195 & 2.097 & 0.0 & 12.85 \\ \ 
    $|E|$ & 0.108 & 3.490 & 58.45 & 53.611 & 1.0 & 398.0 \\ 
    $|V_{car}|$ & 0.289 & 0.772 & 8.160 & 3.75 & 1.0 & 22.0 \\ 
   $mean(\vec{v})$ & 0.434 & 0.635 & 4.349 & 1.863 & 0.0 & 10.894 \\ 
    \bottomrule
\end{tabular}
\
\end{table}
By evaluating the absolute error in relation to the usual ranges of the corresponding regressed variables, we can establish the considerable significance of our embeddings. For example, the number of cars $|V_{car}|$ can be estimated with a deviation of less than one car on average. Similar results can be observed for the number of edges, velocities of cars and the number of interaction types ($|E_{lon}|$, $|E_{lat}|$, $|E_{int}|$), that were normalized to the number of cars in the regarding scene.

Once we have confirmed the meaningfulness of the information encoded in the embedding space, our focus shifts towards examining the distribution of this information. Of particular interest is the continuity within the embedding space. To visualize this high-dimensional space effectively, we employ dimensionality reduction techniques, specifically PCA, to reduce the dimensionality down to two dimensions. Subsequently, we assign colours to the samples based on certain handcrafted features previously utilized in the aforementioned step (see \Cref{tab:hand_crafted_features_regression}).

\begin{figure}[t]
    \centering
    \begin{minipage}[t]{0.49\linewidth}
    \includegraphics[width=\linewidth]{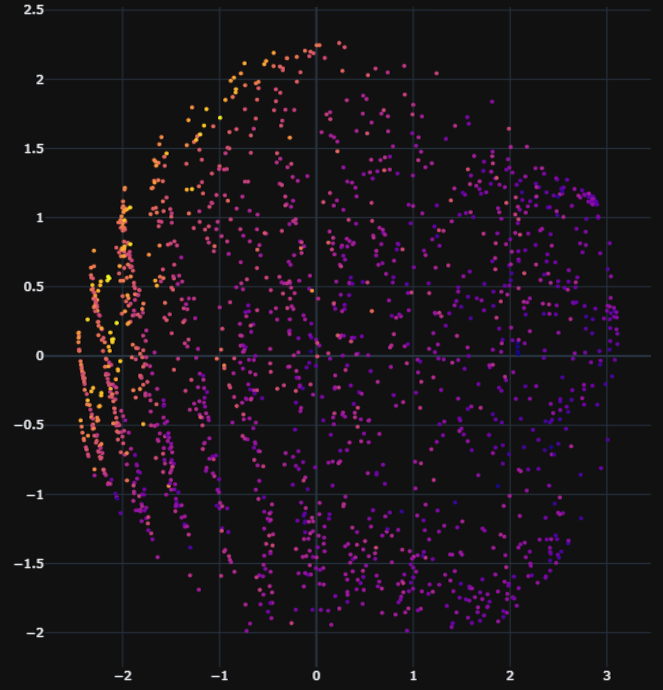}
    \end{minipage}
    \begin{minipage}[t]{0.49\linewidth}
    \includegraphics[width=\linewidth]{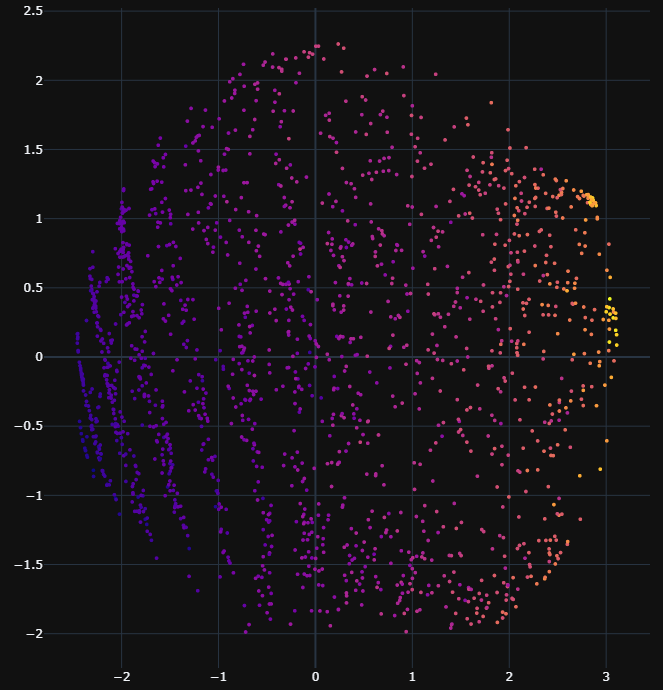}
    \end{minipage}
    \caption{The PCA-reduced embedding space coloured according to the average velocity \gradientbox{blue}{red!60}{0\,m/s ...}\gradientbox{red!60}{yellow}{11\,m/s} of the cars present in the scene (left) and according to the number of vehicles  \gradientbox{blue}{red!60}{1 ...}\gradientbox{red!60}{yellow}{22} (right).}
    \label{fig:pca_plots}
\end{figure}

The first notable example can be seen in \Cref{fig:pca_plots} (left), where the average speed of cars ($mean(\vec{v})$) is coloured. A gradient from top-left to bottom-right is observable, from high speed to low speed. This indicates that some scene-describing continuity is preserved by the embedding space.

The second important example is observable by colouring the number of cars $|V_{car}|$ that are present in the scene. This can be seen in \Cref{fig:pca_plots} (right). Especially on the left-hand side of the figure, around six clearly separated line shaped clusters are visible. Just like before, a gradient is observable. This time from left to right. What is interesting is that those lines encode the number of cars in the scene in an ordinal fashion. The left most line contains mostly scenes with one car, and with every line, one more car joins the scenes.
After six lines, this pattern seems to disappear. The reason for this could be the greater number of cars in scenes, which leads to the possibility of more intricate interactions. In order to distinguish between these scenes, the model would require a more sophisticated approach. 
When observing further features, similar observations could be made.

In concluding our examination of the embedding space, we proceed to test and observe three significant properties of our approach. Firstly, we verify that similar graphs exhibit closer proximity to one another compared to dissimilar graphs. Secondly, we establish that our fixed-dimensional embedding space encompasses important information that can be effectively leveraged in subsequent tasks. Lastly, we confirm that the embedding space demonstrates continuity in at least some of the traffic scene defining information.

\subsection{Clustering}

\begin{figure}[t]
    \centering
    \begin{minipage}[t]{0.32\linewidth}
    \includegraphics[width=\linewidth]{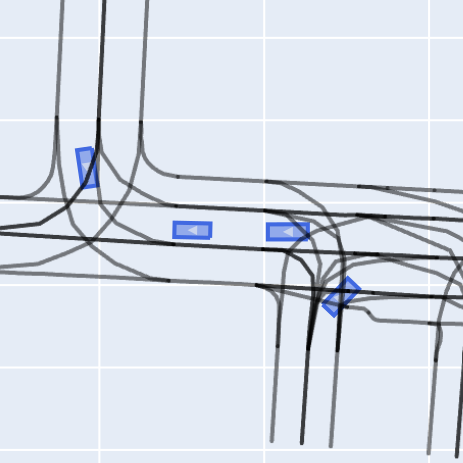}
    \end{minipage}
    \begin{minipage}[t]{0.32\linewidth}
    \includegraphics[width=\linewidth]{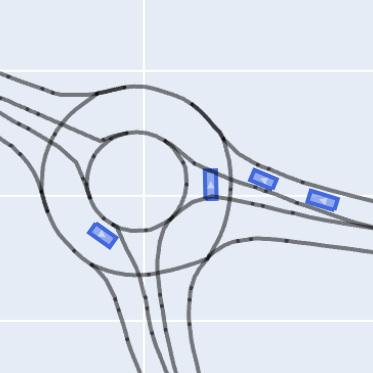}
    \end{minipage}
    \begin{minipage}[t]{0.32\linewidth}
    \includegraphics[width=\linewidth]{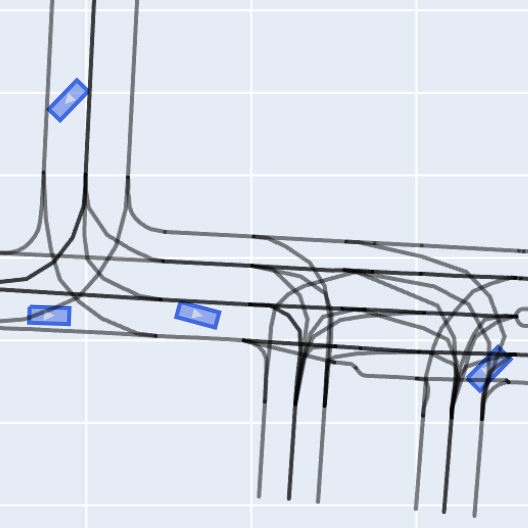}
    \end{minipage}\\
    \vspace{0.5ex}
    \hrule
    \vspace{0.5ex}
    \begin{minipage}[t]{0.32\linewidth}
    \includegraphics[width=\linewidth]{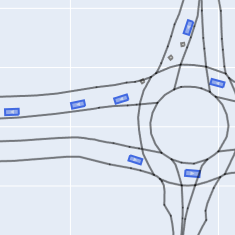}
    \end{minipage}
    \begin{minipage}[t]{0.32\linewidth}
    \includegraphics[width=\linewidth]{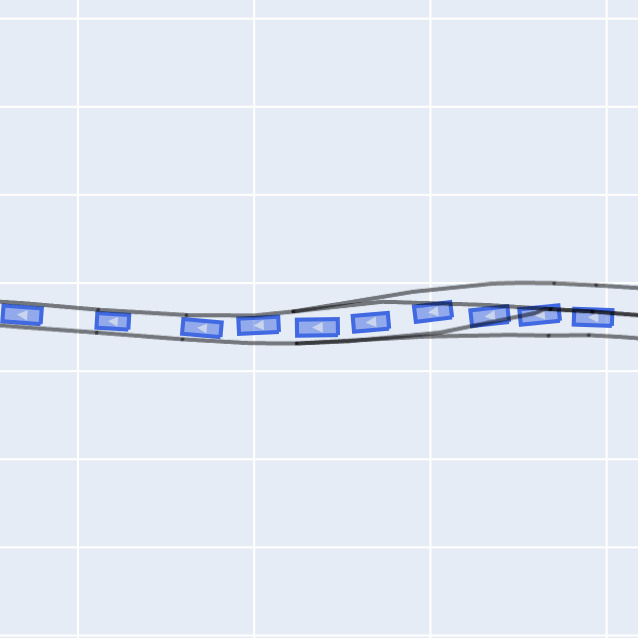}
    \end{minipage}
    \begin{minipage}[t]{0.32\linewidth}
    \includegraphics[width=\linewidth]{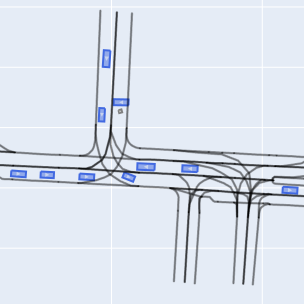}
    \end{minipage}
    \caption{Three examples of traffic scenes, each of two different clusters (top/bottom) from \Cref{fig:umap_clusters}}
    \label{fig:cluster_example_traffic_scenes}
\end{figure}

\begin{figure}
    \centering\includegraphics[width=0.9\columnwidth]{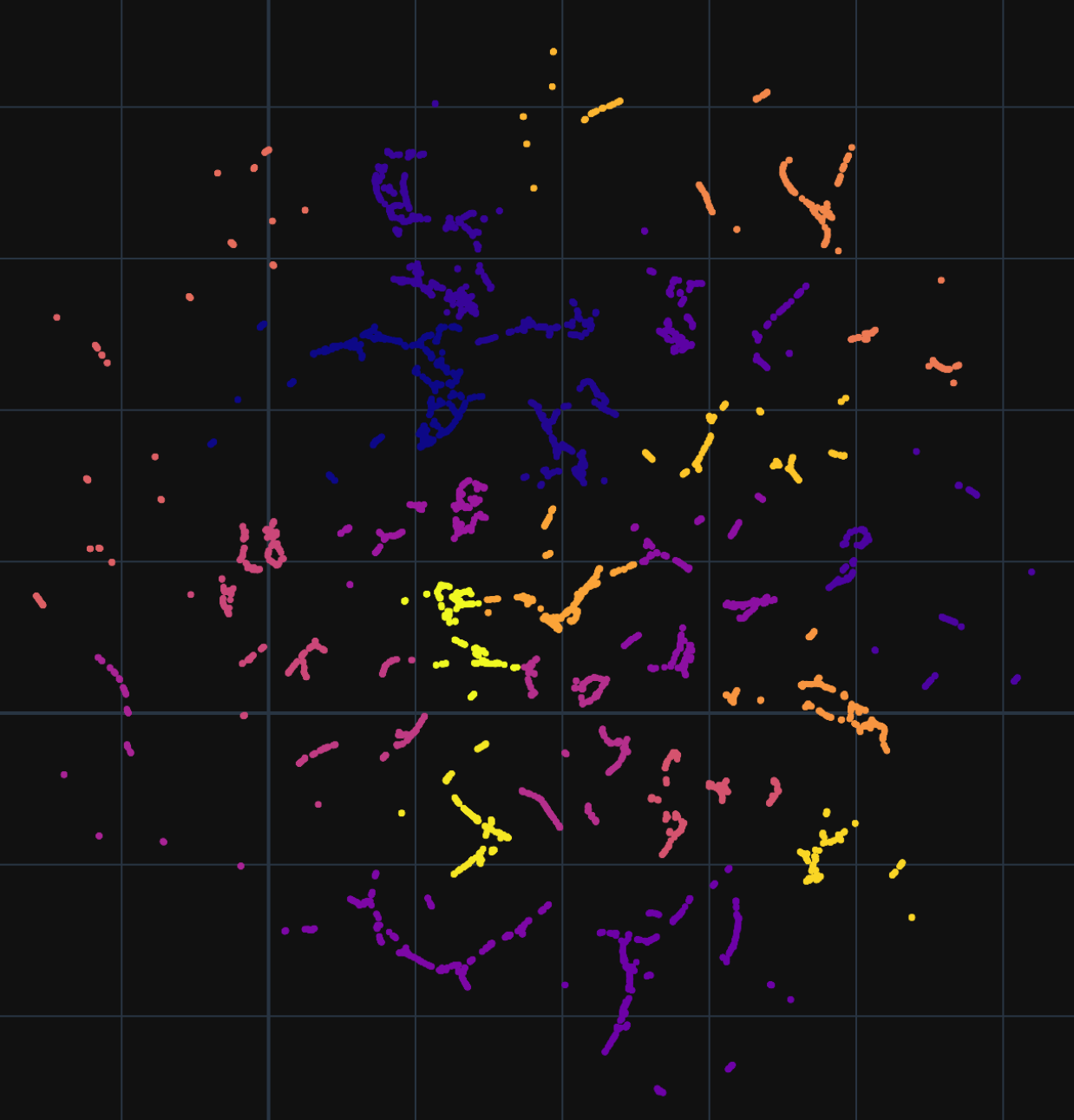}
    \caption{The embedding space $S$ was dimensionally reduced to a two-dimensional space using the UMAP algorithm. Hierarchical clustering was employed to assign data points to 25 distinct clusters, which are represented by different colours.}
    \label{fig:umap_clusters}
\end{figure}

To qualitatively assess the encodings on a deeper semantic level, we apply clustering of the test dataset.
Firstly, dimensionality reduction is performed using the UMAP algorithm \cite{mcinnes_umap_2020} ($n\_neighbors=5, min\_dist=0.0, n\_components=2$). On the resulting two-dimensional space, agglomerative hierarchical clustering is executed (see \Cref{fig:umap_clusters}). The number of clusters is determined by testing a reasonable range of clusters ($[2, 25]$) against their silhouette score and picking the version with the highest score. Generally, we notice that as the number of clusters increases, the silhouette score also increases until approximately 250 clusters, reaching a score of approximately 0.75. After that point, the silhouette score starts to decrease. In order to still being able to evaluate the clusters by hand, we capped the number of clusters at 25 with a silhouette score of 0.617.

\Cref{fig:cluster_example_traffic_scenes} displays six traffic scenes chosen from two distinct clusters of \Cref{fig:umap_clusters}. The top three scenes from the first cluster show a small group of two traffic participants driving behind each other, some individual participants in an intersecting lane, in addition to one vehicle that does not interact with any other vehicles. It is noteworthy, that our approach seems to not only capture similar scenes for the same road type, but transcends road types to even make scenes on different roads comparable.
The traffic scenes in the second cluster (\Cref{fig:cluster_example_traffic_scenes} bottom) exhibit long queues of traffic participants.
In line with the aforementioned clusters, the remaining clusters generated exhibit analogous patterns within their traffic constellations. In general, clusters depicting similar traffic scenarios were discovered across diverse road configurations in various locations.

This qualitative analysis serves as an illustration that the model has successfully acquired a semantic representation of the traffic scenes.

\section{Conclusion and Outlook}
\label{sec:conclusion}

We show through quantitative and qualitative analysis, that Graph Neural Networks combined with contrastive learning are a viable approach for encoding complex traffic scenes and making them comparable, without relying on handcrafted labels and features, but instead build on the underlying interaction graph structure between vehicles.
More specifically, we show the following different properties:

Firstly we show, that the resulting embedding space has generalized to discriminate positive and negative samples from an anchor.

Secondly we show that our embedding space encodes important graph level properties that could be useful for downstream tasks.

By qualitatively analysing formed clusters, we show that our approach can even identify similar traffic scenes on completely different street types successfully.

Those properties make our graph encoding a candidate to employ in further downstream tasks.
One application could be the use of generative models. Often times, generative models such as autoencoders necessitate a direct comparison between samples to compute a loss. With our encoder, the loss could simply be the Euclidean distance between graph encodings.
Another application is in testing of highly automated driving functions. Embeddings like ours have various applications. For instance, they can help in identifying gaps in training sets. Additionally, they can generate new scenarios by swapping street types or intersections while maintaining similarity to existing ones.

\section{Acknowledgement}
The research leading to these results is funded by the German Federal Ministry for Economic Affairs and Climate Action" within the project “Verifikations- und Validierungsmethoden automatisierter Fahrzeuge im urbanen Umfeld". The project is a part of the PEGASUS family. The authors would like to thank the consortium for the successful cooperation.

\bibliographystyle{IEEEtran}

\bibliography{references_clean}

\end{document}